\documentclass[runningheads]{llncs}

\usepackage[T1]{fontenc}
\usepackage{amsmath,amssymb,amsfonts}
\usepackage{array}
\usepackage{graphicx}
\usepackage{makecell}
\usepackage{verbatim}
\usepackage{url}
\usepackage{subcaption}

\usepackage[hidelinks]{hyperref}
\newcommand{\cvmetric}[2]{\makecell[c]{#1\\{\scriptsize $\pm$ #2}}}

\begin{document}

\title{Interpretable Depression Detection from Social Media Text Using LLM-Derived Embeddings}
\titlerunning{Interpretable Depression Detection from Social Media Text}

\author{Samuel Kim\inst{1} \and Oghenemaro Imieye\inst{1} \and Yunting Yin\inst{2}\textsuperscript{*}}
\authorrunning{S. Kim et al.}
\institute{Department of Computer Science, Earlham College, Richmond, IN, USA\\
\email{skim24@earlham.edu; ooimieye23@earlham.edu}
\and
Department of Computer Science, Eastern Michigan University, Ypsilanti, MI, USA\\
\email{yyin@emich.edu}\\
\textsuperscript{*}Corresponding author}

\maketitle

\begin{abstract}
Accurate and interpretable detection of depressive language in social media can support early identification of mental health conditions and inform timely interventions. In this paper, we investigate the use of large language models (LLMs) and traditional machine learning classifiers for three social media-based mental health prediction tasks: binary depression classification, depression severity classification, and differential diagnosis among depression, PTSD, and anxiety. We compare zero-shot LLMs with supervised classifiers trained on conventional text embeddings, psycholinguistic features, and embeddings derived from LLM-generated mental health summaries. Across multiple publicly available social media text datasets and five-fold cross-validation experiments, we find that zero-shot LLMs exhibit strong performance and generalization in binary depression classification, but struggle with fine-grained severity prediction. In contrast, supervised models trained on LLM summary embeddings often achieve more accurate and consistent performance, particularly for multi-class and ordinal classification tasks. These findings highlight both the strengths and limitations of current LLMs for mental health prediction and suggest that using LLMs as semantic interpreters, rather than solely as end-to-end classifiers, may provide a promising direction for building more effective and interpretable mental health assessment systems.

\end{abstract}

\section{Introduction}
Mental health disorders such as depression affect hundreds of millions of individuals worldwide, with many cases remaining undiagnosed or untreated due to social stigma, cost, or lack of access to care. As individuals increasingly express their thoughts and emotions on social media, these platforms have become a valuable source of real-time data for assessing psychological well-being. Automatic detection of depressive language from social media posts can be a promising tool for large-scale, low-cost mental health screening and intervention. Prior approaches to depression classification have typically relied on two types of features: psycholinguistic markers and text embeddings derived from pretrained language models. While psycholinguistic features such as those derived from the Linguistic Inquiry and Word Count (LIWC) lexicon offer interpretability, they are limited in expressiveness. On the other hand, traditional sentence embeddings capture rich semantic information but may lack the specific affective cues that are critical for mental health prediction tasks.

In this work, we propose a novel prompt-based embedding approach that makes use of the reasoning capabilities of LLMs to produce embeddings with more interpretability and semantic richness. Instead of directly embedding the raw input text, we prompt an LLM with a mental health-oriented question and feed it with users' social media post. We then extract embedding from this LLM summary using a sentence encoder and use it as input to classifiers. This method induces reasoning by forcing the LLM to generate a semantically enriched interpretation beyond surface-level syntax. It also reduces noise by allowing the LLM to filter irrelevant information. Moreover, it enhances interpretability by producing intermediate summaries that can be shown to clinicians as part of an intervention or diagnosis. We evaluate our approach on five social media-based depression datasets and find that LLM-derived summary embeddings, while not consistently superior, often enhance predictive performance over models using raw text embeddings in supervised tasks, and outperform a pre-trained large language model in ordinal multi-class classification.

The remainder of this paper is organized as follows. Section~\ref{section2} presents a review of related work, including traditional text-based approaches for mental health prediction from social media and recent applications of large language models in this domain. Section~\ref{section3} introduces our methodology, including data preprocessing, feature extraction using text embeddings and psycholinguistic features, and the generation of LLM-based summary embeddings. Section~\ref{section4} reports experimental results across three classification tasks: binary depression classification, depression severity classification, and differential diagnosis among depression, anxiety, and PTSD. Finally, Section~\ref{section5} concludes the paper with a discussion of key findings and future directions.

\section{Related Works}
\label{section2}

\subsection{Language-Based Mental Health Assessment}
Textual data, whether derived from written language, transcribed speech, or online interactions, offers a powerful lens into mental health, with numerous studies demonstrating that linguistic patterns can reflect emotional states and clinical symptoms associated with mental disorders. Stress detection has been explored using a range of textual sources, including online blog and forum posts \cite{b1,b2,b3} and social media interactions \cite{b4,b5,b6,b7}. PTSD diagnosis has also been performed using clinical patient narratives \cite{b8,b9}, online surveys \cite{b10}, and transcribed voicemails \cite{b11}. Sawalha et al. \cite{b12} argue that sentiment analysis of transcribed text from semi-structured virtual interviews can effectively identify individuals with PTSD using a Random Forest model with VADER sentiment scores. Mansoor et al. \cite{b14} introduce a multimodal AI model that analyzes multilingual social media data to detect early signs of mental health crises, and emphasize the need for ethical safeguards and culturally sensitive applications in real-world mental health systems. Althoff et al. \cite{b15} present a large-scale quantitative analysis of text-message-based counseling conversations using computational discourse methods. Ewbank et al. \cite{b16} develop a deep learning model to automatically categorize patient utterances during internet-enabled cognitive behavioral therapy. Bantilan et al. \cite{b17} propose an NLP model to detect suicide risk in patient messages during teletherapy, using therapist intervention patterns and expert annotations to label risk levels. These studies highlight the growing potential of natural language processing (NLP) for mental health assessment and intervention, and show the importance of contextual and linguistic features in the real-world applicability of such models.

\subsection{Large Language Models for Mental Health Prediction}
Recent advancements in large language models have enabled their application across a wide range of domains, including the analysis and prediction of mental health conditions. Xu et al. \cite{b50} evaluates the performance of several large language models on mental health prediction tasks using online text data. Their findings suggest that while zero- and few-shot prompting yield limited results, instruction fine-tuning significantly boosts accuracy. Boggavarapu et al. \cite{b51} explore the use of LLMs enhanced with Retrieval-Augmented Generation to predict mental health-related ICD-10-CM codes from clinical notes, and find that current LLMs still struggle with accurately interpreting these complex codes. Their findings suggest the need for better integration of structured medical knowledge into these models. Malgaroli et al. \cite{b52} discuss the potential of LLMs to advance mental health care through improved diagnostics, monitoring, and treatment. They also identify challenges such as bias, accessibility, and data representation. Qian el al. \cite{b53} explore how foundation models, such as LLMs, are transforming digital mental health through personalized diagnostics, real-time monitoring, emotion recognition, and adaptive interventions using multimodal data. They propose a sociotechnical framework that integrates brain-inspired AI, and clinical oversight with ethical considerations. Hua et al. \cite{b70} review the current landscape of LLM applications in mental health care, and conclude that there is promising use cases in counseling and clinical support, but most studies lack standardized evaluation methods. Together, these studies show the growing potential of LLMs in mental health diagnosis and care. Addressing the challenges in model reliability, interpretability, and evaluation rigor will be crucial for the integration of LLMs into real-world clinical settings.

\subsection{LLMs as Reasoning and Summarization Tools}
Beyond direct classification, large language models can reason over context and summarize complex narratives to generate structured representations that support downstream machine learning tasks. SemCSE \cite{b81} was proposed as an unsupervised embedding method that uses LLM-generated summaries of scientific abstracts as semantically enriched training signals, showing that summary-based representations can improve semantic embedding quality for downstream tasks. Liu et al. \cite{b82} find that embeddings derived from LLM-generated descriptions of genes and cell metadata can serve as effective semantic representations to improving performance on downstream biological analysis tasks. Moreover, LLMs can be used as intermediate representation generators. Li and Zhou propose a training-free embedding approach \cite{b83} that combines mixture-of-experts routing weights with hidden-state representations, showing that internal LLM activations can also provide useful semantic representations for downstream embedding tasks. These works motivate our exploration of summary-based embeddings as a mechanism for combining the semantic reasoning capabilities of LLMs with the transparency and efficiency of traditional classifiers.

\section{Methods}
\label{section3}
Our methodology is organized into five main stages: data preprocessing, text embedding and psycholinguistic feature extraction, zero-shot LLM prompting, LLM summary embedding generation, and model training and evaluation.

\subsection{Data Preprocessing}
We preprocess five publicly available social media-based mental health datasets for our experiments: MHB \cite{b100}, CAMS \cite{b200}, HelaDepDet \cite{b300}, RMHD \cite{b400}, and DepressionEmo \cite{b500}. Each dataset consists of short, user-generated text entries annotated with mental health labels, primarily related to depression. To enable evaluation of the model's ability to distinguish between depressive and non-depressive language, we include two general-domain Reddit control datasets, RedditAITA \cite{b600} and RedditTIFU \cite{b800}, as non-depression examples in the combined dataset. These two datasets were selected because they are publicly available corpora containing informal first-person narratives that are stylistically similar to the social media posts found in the mental health datasets. To maintain a balanced representation between the two control sources, RedditTIFU was sampled to contain a comparable number of instances as RedditAITA.

\begin{table}[ht]
\centering
\caption{Statistics of Datasets Used Across Classification Tasks}
\label{tab:dataset_stats}
\footnotesize
\setlength{\tabcolsep}{4pt}
\renewcommand{\arraystretch}{1.22}
\begin{tabular}{|p{0.17\textwidth}|p{0.13\textwidth}|>{\centering\arraybackslash}p{0.10\textwidth}|>{\centering\arraybackslash}p{0.10\textwidth}|p{0.34\textwidth}|}
\hline
\textbf{Dataset} & \textbf{Task} & \makecell[c]{\textbf{Size}\\\textbf{(posts)}} &
\makecell[c]{\textbf{Avg.}\\\textbf{words}} &
\makecell[l]{\textbf{Labels}\\\textbf{(\%)}} \\
\hline
\makecell[l]{MHB\\\cite{b100}} & \makecell[l]{Binary\\Differential} & 7,452 & 253 & \makecell[l]{Depression: 43\\Anxiety: 45\\PTSD: 12} \\
\hline
\makecell[l]{CAMS\\\cite{b200}} & \makecell[l]{Binary} & 4,042 & 179 & \makecell[l]{All depression} \\
\hline
\makecell[l]{HelaDepDet\\\cite{b300}} & \makecell[l]{Binary\\Severity} & 33,498 & 120 & \makecell[l]{Minimum: 25\\Mild: 25\\Moderate: 23\\Severe: 27} \\
\hline
\makecell[l]{RMHD\\\cite{b400}} & \makecell[l]{Binary\\Differential} & 658 & 236 & \makecell[l]{Depression: 34\\Anxiety: 25\\Other: 41} \\
\hline
\makecell[l]{Depression\\Emo\\\cite{b500}} & \makecell[l]{Binary} & 4,830 & 95 & \makecell[l]{All depression} \\
\hline
\makecell[l]{RedditAITA\\\cite{b600}} & \makecell[l]{Binary} & 24,795 & 386 & \makecell[l]{All non-depression control} \\
\hline
\makecell[l]{RedditTIFU\\\cite{b800}} & \makecell[l]{Binary} & 25,685 & 341 & \makecell[l]{All non-depression control} \\
\hline
\textbf{Total (unique)} & --- & 100,960 & --- & --- \\
\hline
\end{tabular}
\end{table}

We begin by cleaning each dataset through a series of preprocessing steps. Duplicate entries are removed, and we retain only posts with text lengths falling between the 10th and 90th percentiles to exclude outliers. For each dataset, only the columns relevant to the downstream tasks are preserved, while metadata fields such as text length, number of posts by the author, and number of likes are removed. Our experimental setup supports three classification tasks:

\begin{enumerate}
\item \textbf{Binary Depression Classification:} We combine all five depression-related datasets with RedditAITA and RedditTIFU non-depression controls to train models that distinguish between depressive and non-depressive content.
\item \textbf{Depression Severity Classification:} Using the HelaDepDet \cite{b300} dataset, which provides graded depression severity labels, we train models to predict levels including minimum, mild, moderate, and severe.
\item \textbf{Differential Diagnosis Classification:} We use the MHB \cite{b100} and RMHD \cite{b400} datasets, which contain multi-class annotations for depression, anxiety, and PTSD, to assess the model's ability to differentiate between related mental health conditions.
\end{enumerate}

Descriptive statistics for the preprocessed datasets, including the number of posts, label categories, and average text length measured in words, are summarized in Table~\ref{tab:dataset_stats}. For supervised experiments using text embeddings and machine learning classifiers, we applied deterministic five-fold stratified cross-validation. Each task dataset was partitioned into five folds while preserving the class distribution within each fold. During each iteration, models were trained on four folds and evaluated on the remaining held-out fold. Notably, two zero-shot LLM-based classification approaches do not require model training. These methods were evaluated directly on the full task datasets using prompting strategies without parameter updates.

\subsection{Text Embedding and Psycholinguistic Feature Extraction}
To establish baseline performance, we evaluate classifiers trained on the combination of two types of traditional feature representations:

\subsubsection{Text Embeddings}
We extract contextualized sentence-level embeddings using the \texttt{all-mpnet-base-v2} \cite{b700} model from the SentenceTransformers library. Each social media post is passed through the pretrained model to obtain a 768-dimensional fixed-size embedding vector.

\subsubsection{Psycholinguistic Features}
We compute psycholinguistic features using the Linguistic Inquiry and Word Count (LIWC) lexicon. Each post is analyzed to yield normalized frequencies of relevant categories, including affective processes, cognitive processes, and pronoun usage. The resulting feature vectors are standardized using z-score normalization.

\subsubsection{Classification Models}
\label{classification_models}
We evaluate three traditional machine learning classifiers using both text embeddings alone and the concatenation of text embeddings with psycholinguistic features. Logistic Regression was configured with L2 regularization ($C=1.0$) and a maximum of 1,000 iterations. The Support Vector Machine (SVM) employed a linear kernel with $C=1.0$, while the Random Forest classifier consisted of 100 decision trees. All supervised experiments were conducted using deterministic five-fold stratified cross-validation. Performance was evaluated using accuracy, precision, recall, macro-F1, and class-wise F1 where applicable, and we report the mean and standard deviation across the five folds.

\subsection{Zero-Shot LLM-Based Classification}

To evaluate large language models as direct zero-shot classifiers, we use the OpenAI GPT-4o and GPT-5.5 API. For each social media post, we send a prompt requesting a binary or multi-class classification label, depending on the task setting. The prompt format for the binary classification task is as follows:
\begin{quote}
You are a mental health expert. Read the following social media post and determine the user's mental health condition. Choose from the following labels: Depression, Non-depression.
\end{quote}

For severity detection and differential diagnosis, the label options are modified accordingly. The model's textual response is parsed as the predicted label. No training or fine-tuning is applied.

\subsection{Prompted LLM Summary Embedding}
In this approach, we prompt GPT-5.5, the best-performing LLM in preliminary experiments, to interpret and summarize the user's mental state based on the content of a social media post. The goal is to produce a concise, clinically oriented description that captures signals relevant to mental health assessment. The prompt format for the LLM summary generation is as follows:

\begin{quote}
You are a mental health expert. Read the following social media post and describe the user's mental state in one or two sentences. Focus on emotional tone, cognitive state, and any signs of mental health conditions. Avoid quoting the post verbatim.
\end{quote}

We embed the LLM-generated response using the \texttt{all-mpnet-base-v2} \cite{b700} sentence embedding model, resulting in a 768-dimensional feature vector. This vector captures task-relevant affective semantics abstracted from the original text. We then train the same set of classifiers described in Section~\ref{classification_models} on these embeddings to assess whether LLM-generated paraphrased representations improve predictive performance.

An additional advantage of this approach is interpretability. Traditional sentence embeddings encode semantic information in a high-dimensional latent space that is difficult for humans to inspect directly. With an intermediate summary, the generated prediction by downstream classifiers has a transparent representation that can be reviewed by clinicians or other stakeholders to better understand the information being used.

\subsection{Evaluation Metrics}
Results are reported as the mean and standard deviation across five-fold cross-validation. For the binary depression classification task, we report accuracy, precision, recall, and F1-score. For the depression severity detection and differential diagnosis tasks, we report accuracy, macro F1, and class-wise F1-scores to provide a more comprehensive assessment of performance across categories. All metrics are computed using the \texttt{scikit-learn} \cite{b801} library.

\begin{figure}[htbp]
\centering
\includegraphics[width=\textwidth]{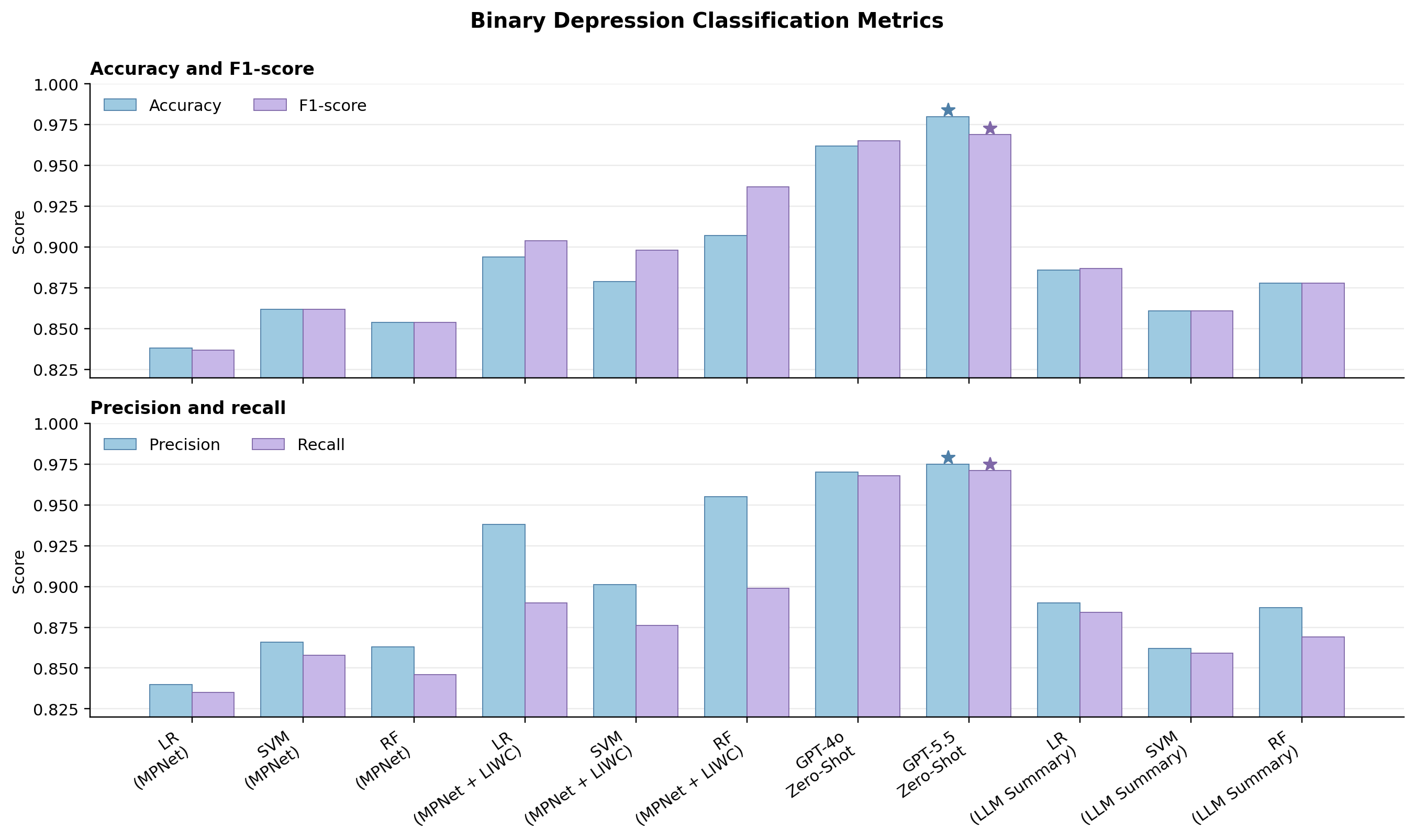}
\caption{Binary classification accuracy and F1-score across feature representations and classifiers. Error bars show cross-validation standard deviation.}
\label{fig:task1_metrics_updated}
\end{figure}

\section{Experiment Results}
\label{section4}

In this section, we present the evaluation results across three mental health classification tasks. Each task is designed to assess the model's ability to handle different types of diagnoses. We begin with binary depression classification, where the goal is to distinguish between depressive and non-depressive content using a combination of datasets. Next, we examine depression severity classification, in which the models predict fine-grained severity levels using the HelaDepDet \cite{b300} dataset. Finally, we evaluate the performance of the models and embeddings on the task of differential diagnosis classification among depression, anxiety, and PTSD, using the MHB \cite{b100} and RMHD \cite{b400} datasets. 

\begin{table}[ht]
\centering
\caption{Binary classification metrics by model and feature type. Cross-validation rows report five-fold mean $\pm$ standard deviation.}
\label{tab:task1_metrics}
\footnotesize
\setlength{\tabcolsep}{3pt}
\renewcommand{\arraystretch}{1.2}
\begin{tabular}{|p{0.38\textwidth}|*{4}{>{\centering\arraybackslash}p{0.125\textwidth}|}}
\hline
\textbf{Model} & \textbf{Accuracy} & \textbf{Precision} & \textbf{Recall} & \textbf{F1-score} \\
\hline
\makecell[l]{Logistic Regression\\(MPNet)} &
\cvmetric{0.838}{0.001} &
\cvmetric{0.840}{0.001} &
\cvmetric{0.835}{0.001} &
\cvmetric{0.837}{0.002} \\
\hline
\makecell[l]{SVM\\(MPNet)} &
\cvmetric{0.862}{0.003} &
\cvmetric{0.866}{0.002} &
\cvmetric{0.858}{0.003} &
\cvmetric{0.862}{0.003} \\
\hline
\makecell[l]{Random Forest\\(MPNet)} &
\cvmetric{0.854}{0.002} &
\cvmetric{0.863}{0.002} &
\cvmetric{0.846}{0.003} &
\cvmetric{0.854}{0.002} \\
\hline
\makecell[l]{Logistic Regression\\(MPNet + LIWC)} &
\cvmetric{0.894}{0.002} &
\cvmetric{0.938}{0.001} &
\cvmetric{0.890}{0.002} &
\cvmetric{0.904}{0.002} \\
\hline
\makecell[l]{SVM\\(MPNet + LIWC)} &
\cvmetric{0.879}{0.001} &
\cvmetric{0.901}{0.001} &
\cvmetric{0.876}{0.001} &
\cvmetric{0.898}{0.001} \\
\hline
\makecell[l]{Random Forest\\(MPNet + LIWC)} &
\cvmetric{0.907}{0.002} &
\cvmetric{0.955}{0.001} &
\cvmetric{0.899}{0.002} &
\cvmetric{0.937}{0.002} \\
\hline
\makecell[l]{GPT-4o Zero-Shot} & 0.962 & 0.970 & 0.968 & 0.965 \\
\hline
\makecell[l]{GPT-5.5 Zero-Shot} & \textbf{0.980} & \textbf{0.975} & \textbf{0.971} & \textbf{0.969} \\
\hline
\makecell[l]{Logistic Regression\\(LLM Summary)} &
\cvmetric{0.886}{0.001} &
\cvmetric{0.890}{0.000} &
\cvmetric{0.884}{0.002} &
\cvmetric{0.887}{0.001} \\
\hline
\makecell[l]{SVM\\(LLM Summary)} &
\cvmetric{0.861}{0.000} &
\cvmetric{0.862}{0.001} &
\cvmetric{0.859}{0.001} &
\cvmetric{0.861}{0.000} \\
\hline
\makecell[l]{Random Forest\\(LLM Summary)} &
\cvmetric{0.878}{0.001} &
\cvmetric{0.887}{0.001} &
\cvmetric{0.869}{0.001} &
\cvmetric{0.878}{0.001} \\
\hline
\end{tabular}
\end{table}

\subsection{Binary Depression Classification}

For the binary classification task of determining whether a social media post is depressive or non-depressive, performance metrics are summarized in Table \ref{tab:task1_metrics} and shown in Figure \ref{fig:task1_metrics_updated}. Among all models evaluated, the zero-shot GPT-5.5 classifier achieved the highest overall accuracy, outperforming both traditional machine learning models using psycholinguistic and text-based embeddings, as well as models predicting based on LLM-generated summary embeddings.

\begin{figure}[htbp]
\centering
\includegraphics[width=0.88\textwidth]{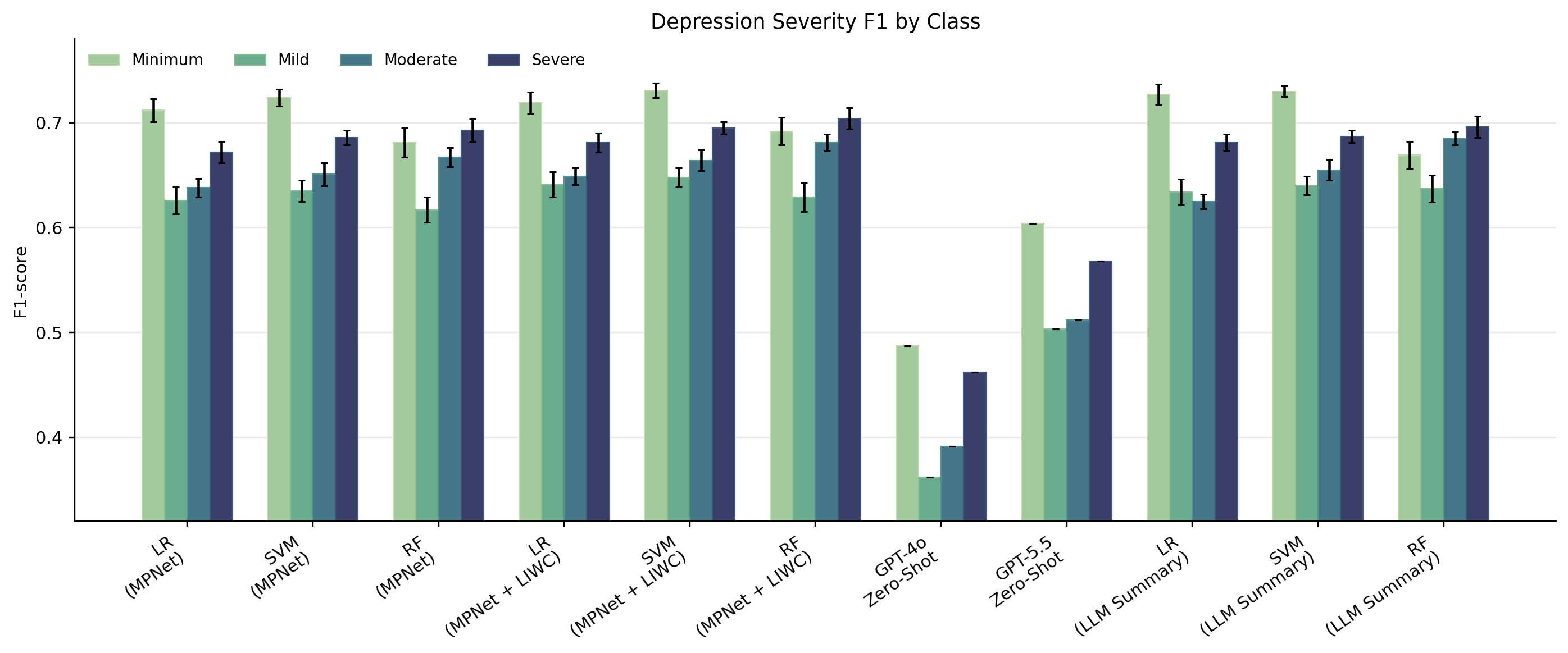}
\caption{Class-wise F1-scores for depression severity classification. Error bars indicate standard deviations across five-fold cross-validation for supervised models.}
\label{fig:task2_f1_updated}
\end{figure}

Notably, the zero-shot LLM was not specifically fine-tuned for depression detection, yet demonstrated strong performance, likely due to its extensive pretraining on large-scale, diverse datasets. This proves the model's impressive generalization capabilities and supports recent findings on the excellent performance of LLMs in zero-shot settings.

Machine learning models using LLM-generated summary embeddings performed better than those using features extracted directly from the raw social media text. This outcome is expected as the summaries provide a condensed, higher-level interpretation of each post, which likely makes implicit depressive cues more accessible.

\subsection{Depression Severity Classification}
\begin{figure}[htbp]
\centering
\begin{subfigure}[b]{0.48\textwidth}
\centering
\includegraphics[width=\textwidth]{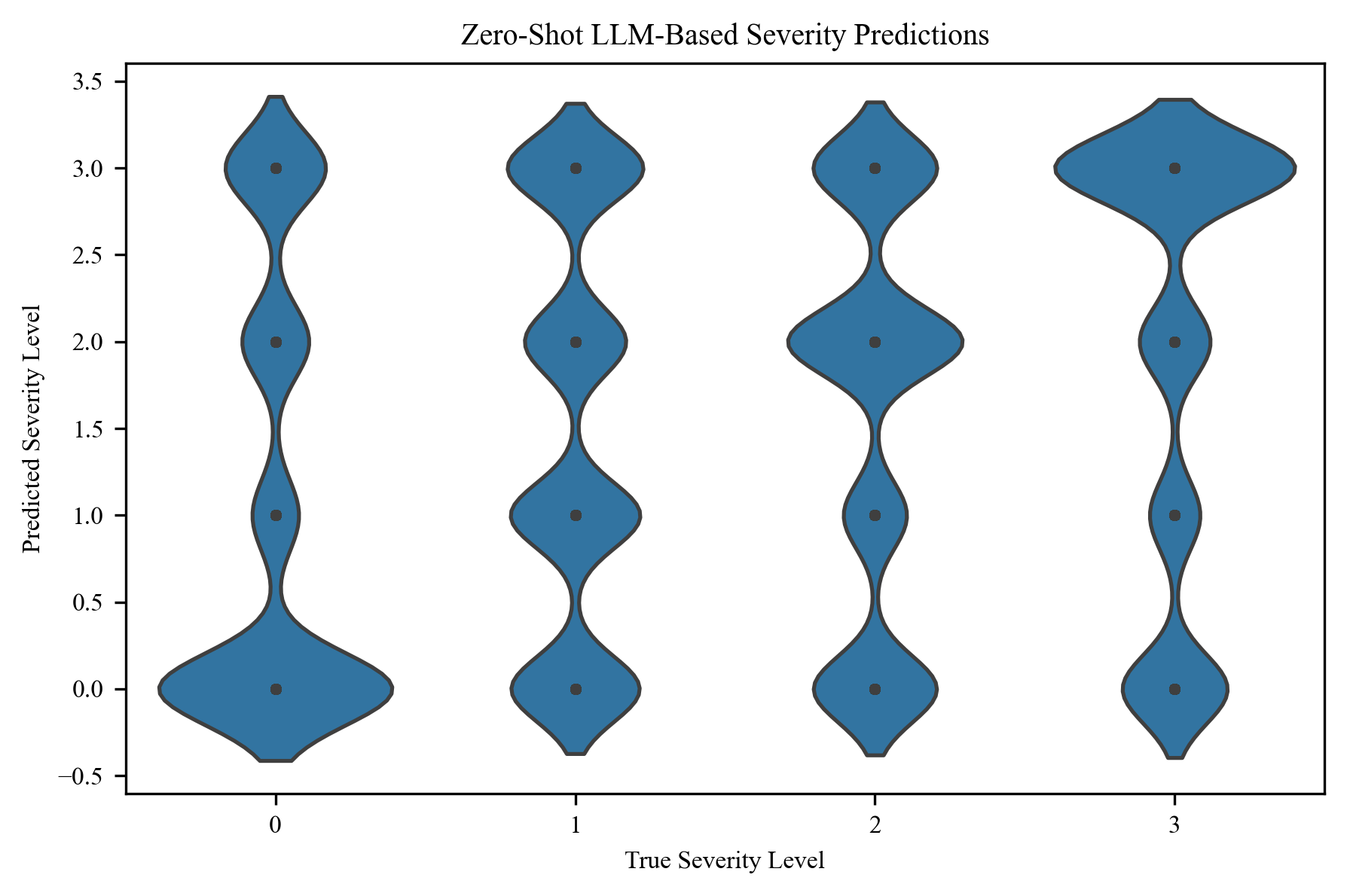}
\label{fig:task2_zero_shot_violin}
\end{subfigure}
\hfill
\begin{subfigure}[b]{0.48\textwidth}
\centering
\includegraphics[width=\textwidth]{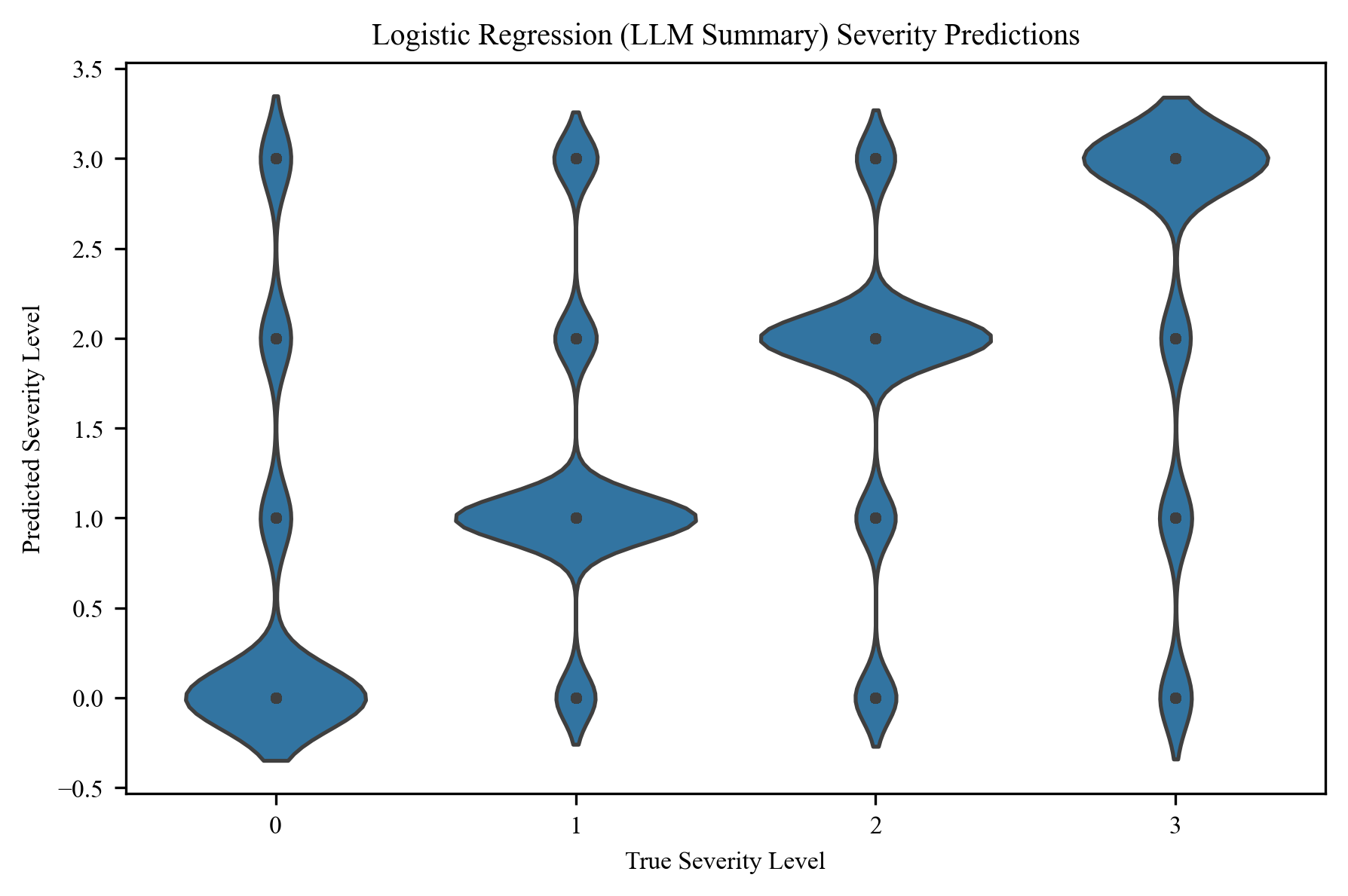}
\label{fig:task2_summary_violin}
\end{subfigure}
\caption{Violin plots of predicted depression severity grouped by the ground-truth, illustrating the separation achieved by a zero-shot LLM (left) and a supervised classifier using LLM-summary embeddings (right).}
\label{fig:task2_violin_comparison}
\end{figure}

To evaluate how well different models and embeddings can assess the severity of depression from users' social media posts, we conducted a multi-class classification task using HelaDepDet \cite{b300} dataset. This dataset includes four ordinal labels representing increasing levels of depression severity, ranging from 0 (minimum) to 3 (severe).

We applied the same classification framework as in the previous task, comparing both zero-shot LLM-based classification and traditional machine learning models with various feature representations. Among all methods, the Logistic Regression classifier using LLM-generated summary embeddings achieved the highest accuracy of 58\%, slightly outperforming models using alternative features, and significantly outperforming the zero-shot LLM-based approach.

Figure \ref{fig:task2_f1_updated} shows the per-class F1 scores for each classifier. Figure \ref{fig:task2_violin_comparison} present violin plots comparing predicted and true severity levels for the worst and best-performing models. We observe that the zero-shot LLM classifier struggles to infer fine-grained severity levels directly from raw text, often failing to reflect the ordinal structure of the labels. In contrast, trained machine learning models benefit from supervised learning, capturing both the semantic and ordinal relationships in the data, and are able to make accurate and consistent depression severity predictions.

\begin{table}[ht]
\centering
\caption{Differential diagnosis classification across feature representations and models}
\label{tab:task3_metrics}
\footnotesize
\setlength{\tabcolsep}{3pt}
\renewcommand{\arraystretch}{1.25}
\begin{tabular}{|p{0.23\textwidth}|*{5}{>{\centering\arraybackslash}p{0.115\textwidth}|}}
\hline
\textbf{Model} & \textbf{Acc.} & \makecell[c]{\textbf{Macro}\\\textbf{F1}} & \makecell[c]{\textbf{Anxiety}\\\textbf{F1}} & \makecell[c]{\textbf{Depress.}\\\textbf{F1}} & \makecell[c]{\textbf{PTSD}\\\textbf{F1}} \\
\hline
\makecell[l]{Logistic Regression\\(MPNet)} & \cvmetric{0.769}{0.004} & \cvmetric{0.765}{0.007} & \cvmetric{0.753}{0.013} & \cvmetric{0.751}{0.006} & \cvmetric{0.791}{0.005} \\
\hline
\makecell[l]{SVM\\(MPNet)} & \cvmetric{0.781}{0.010} & \cvmetric{0.770}{0.002} & \cvmetric{0.749}{0.008} & \cvmetric{0.775}{0.009} & \cvmetric{0.786}{0.003} \\
\hline
\makecell[l]{Random Forest\\(MPNet)} & \cvmetric{0.754}{0.009} & \cvmetric{0.730}{0.012} & \cvmetric{0.674}{0.032} & \cvmetric{0.754}{0.006} & \cvmetric{0.763}{0.010} \\
\hline
\makecell[l]{Logistic Regression\\(MPNet + LIWC)} & \cvmetric{0.776}{0.005} & \cvmetric{0.773}{0.006} & \cvmetric{0.757}{0.009} & \cvmetric{0.769}{0.003} & \cvmetric{0.793}{0.007} \\
\hline
\makecell[l]{SVM\\(MPNet + LIWC)} & \cvmetric{0.802}{0.004} & \cvmetric{0.804}{0.005} & \cvmetric{0.800}{0.003} & \cvmetric{0.799}{0.005} & \cvmetric{0.814}{0.008} \\
\hline
\makecell[l]{Random Forest\\(MPNet + LIWC)} & \cvmetric{0.769}{0.003} & \cvmetric{0.731}{0.008} & \cvmetric{0.691}{0.013} & \cvmetric{0.732}{0.011} & \cvmetric{0.769}{0.002} \\
\hline
\makecell[l]{GPT-4o Zero-Shot} & 0.875 & 0.855 & 0.825 & 0.868 & 0.873 \\
\hline
\makecell[l]{\textbf{GPT--5.5}\\\textbf{Zero-Shot}} & \textbf{0.891} & \textbf{0.876} & \textbf{0.831} & \textbf{0.902} & \textbf{0.895} \\
\hline
\makecell[l]{Logistic Regression\\(LLM Summary)} & \cvmetric{0.785}{0.008} & \cvmetric{0.776}{0.006} & \cvmetric{0.744}{0.010} & \cvmetric{0.781}{0.005} & \cvmetric{0.803}{0.006} \\
\hline
\makecell[l]{SVM\\(LLM Summary)} & \cvmetric{0.826}{0.005} & \cvmetric{0.819}{0.005} & \cvmetric{0.823}{0.011} & \cvmetric{0.794}{0.002} & \cvmetric{0.840}{0.003} \\
\hline
\makecell[l]{Random Forest\\(LLM Summary)} & \cvmetric{0.772}{0.004} & \cvmetric{0.744}{0.007} & \cvmetric{0.720}{0.008} & \cvmetric{0.734}{0.009} & \cvmetric{0.777}{0.003} \\
\hline
\end{tabular}
\end{table}

\subsection{Differential Diagnosis Classification}
We evaluated model performance on the task of differential diagnosis classification using the MHB \cite{b100} and RMHD \cite{b400} datasets, which include multi-class annotations for depression, PTSD, and anxiety. This task assesses the models' ability to distinguish between related but clinically distinct mental health conditions based on social media text.

Among all evaluated methods, the zero-shot GPT-5.5 and GPT-4o classifiers achieved the highest overall accuracies, slightly outperforming the summary-based embedding approaches, while traditional machine learning models trained on text embeddings combined with LIWC features generally produced the lowest overall accuracies among the evaluated methods. Table~\ref{tab:task3_metrics} shows the F1 score distribution across all classifiers and diagnostic classes. We observe that that most classifiers tend to confuse depression with anxiety, which is expected as the two mental health conditions have overlapping linguistic and emotional patterns, and social media expressions can reflect that. Moreover, PTSD is more consistently distinguished, possibly due to more specific symptom language, such as references to trauma, that sets it apart from the other two.

\section{Conclusion and Future Work}
\label{section5}
This study presents a comparative evaluation of zero-shot LLMs and traditional machine learning models in depression classification tasks based on social media data. We find that zero-shot LLMs have strong performance in binary depression classification, proving their ability to generalize from pretrained knowledge. However, their performance declines in tasks such as severity prediction, where supervised models with LLM-generated summary embeddings show more accurante and consistent performance.

Our evaluation results suggest that LLMs are powerful for mental health prediction tasks, and their contextual summaries are helpful to derive better features. Summary embeddings derived from LLMs capture important semantic cues and can improve traditional models to make more accurate and consistent predictions. Summaries can also serve as human-readable explanations of the signals contributing to a prediction and may help stakeholders assess whether a model's conclusions are supported by meaningful linguistic evidence.  These findings prove the potential of hybrid approaches that combine the generalization capabilities of LLMs with lightweight, interpretable classifiers trained on curated features. Further performance gains may be achieved by exploring advanced prompting strategies, applying few-shot learning, and fine-tuning LLMs on domain-specific mental health data.

\begin{credits}
\subsubsection{\ackname} This research was supported by the Lemann Student/Faculty Collaborative Research Fund at Earlham College. We gratefully acknowledge their funding and support.

\subsubsection{\discintname} The authors have no competing interests to declare that are relevant to the content of this article.
\end{credits}

\end{document}